\pdfoutput=1

\documentclass{article}
\usepackage{amsmath}
\usepackage{subfigure}
\usepackage[dvipsnames]{xcolor}
\usepackage{graphicx}
\PassOptionsToPackage{numbers, compress}{natbib}

\usepackage{lipsum}

\usepackage[final]{neurips_2022}

\usepackage[utf8]{inputenc} 
\usepackage[T1]{fontenc}    
\usepackage[hidelinks]{hyperref}       
\usepackage{url}            
\usepackage{booktabs}       
\usepackage{amsfonts}       
\usepackage{nicefrac}       
\usepackage{microtype}      
\usepackage{xcolor}         
\usepackage{footmisc}
\usepackage{tablefootnote}
\usepackage[export]{adjustbox}

\usepackage{tabularx} 

\newcommand{\Ac}{A_c}

\newcommand{\dregc}{s}
\newcommand{\Ao}{A_\dregc}

\title{New Frontiers in Graph Autoencoders: \\ Joint Community Detection and Link Prediction}

\author{%
  Guillaume Salha-Galvan\thanks{Corresponding author at: \texttt{research@deezer.com}.} \\ Deezer Research \\ Paris, France \And Johannes F. Lutzeyer \\ LIX, École Polytechnique, IP Paris \\ Palaiseau, France \And  George Dasoulas \\ DBMI, Harvard University \\ Cambridge, MA, USA \AND 
  Romain Hennequin \\ Deezer Research  \\ Paris, France \And Michalis Vazirgiannis \\ LIX, École Polytechnique, IP Paris \\ Palaiseau, France
}

\begin{document}

\maketitle

\begin{abstract}
Graph autoencoders (GAE) and variational graph autoencoders (VGAE) emerged as powerful methods for link prediction (LP). Their performances are less impressive on community detection (CD), where they are often outperformed by simpler alternatives such as the Louvain method. It is still unclear to what extent one can improve CD with GAE and VGAE, especially in the absence of node features. It is moreover uncertain whether one could do so while simultaneously preserving good performances on LP in a multi-task setting. In this workshop paper, summarizing results from our journal publication \cite{salhagalvan2022modularity}, we show that jointly addressing these two tasks with high accuracy is possible. For this purpose, we introduce a community-preserving message passing scheme, doping our GAE and VGAE encoders by considering both the initial graph and Louvain-based prior communities when computing embedding spaces. Inspired by modularity-based clustering, we further propose novel training and optimization strategies specifically designed for joint LP and CD. We demonstrate the empirical effectiveness of our approach, referred to as Modularity-Aware GAE and VGAE, on~various~real-world~graphs.
\end{abstract}

\section{Introduction}

Extracting relevant information from nodes of a graph is essential to tackle a wide range of machine learning problems \cite{fortunato2010community,hamilton2020graph,hamilton2017representation,zhang2018network}. This includes \textit{link prediction} (LP) \cite{kumar2020link,liben2007link}, which consists in inferring the presence of new or unobserved edges between node pairs, 
and \textit{community~detection}~(CD) \cite{blondel2008louvain,fortunato2010community}, which consists in clustering nodes into similar groups, according to a chosen similarity metric. 
To address such problems, significant efforts have recently been devoted to the development of \textit{node embedding} methods \cite{hamilton2020graph,hamilton2017representation,kipf2020phd}. These methods aim to learn vectorial representations of nodes in an \textit{embedding space} where node positions should reflect and summarize the initial graph structure. They assess the probability of a new edge between two nodes, or their likelihood of belonging to the same community, by evaluating the proximity of these nodes in the~embedding~space~\cite{choong2018learning,kipf2016-2,wang2017mgae}.

In particular, \textit{graph autoencoders} (GAE) and \textit{variational graph autoencoders} (VGAE) \cite{kipf2016-2,tian2014learning,wang2017mgae,wang2016structural} recently emerged as two powerful families of node embedding methods. Both methods rely on an \textit{encoding-decoding} strategy that consists in \textit{encoding} nodes into an embedding space from which \textit{decoding}, i.e., reconstructing the original graph, should ideally be possible. Originally mainly designed for LP (at least in their modern formulation leveraging \textit{graph neural networks} (GNN)~\cite{kipf2016-2}), the effectiveness of GAE and VGAE models and their extensions on this task has
been experimentally confirmed \cite{grover2019graphite,semiimplicit2019,huang2019rwr,pan2018arga,salha2019-2,aaai20,tran2018multi}. On the other hand, several studies \cite{choong2018learning,choong2020optimizing,salha2021fastgae,salha2019-1} have pointed out their limitations on CD. These studies emphasized that GAEs and VGAEs are often outperformed by simpler CD alternatives, such as the popular Louvain method \cite{blondel2008louvain}. The question of how to improve CD with GAEs and VGAEs remains incompletely addressed, especially in the absence of node features. Moreover, it is still unclear to which extent one can improve CD with these models without simultaneously deteriorating LP, and jointly address these two problems. 
These questions are highly relevant in practice, as learning node embedding spaces suitable for multi-task settings leads to consistent inference between tasks and saves costs in~real-world~applications.


This paper\footnote{\label{footnote:journal} This workshop paper summarizes results from our journal article \textit{``Modularity-Aware Graph Autoencoders for Joint Community Detection and Link Prediction''} accepted for publication in Elsevier's Neural Networks journal in 2022 \cite{salhagalvan2022modularity}. The purpose of our submission to GLFrontiers was to present this work to a live audience.} 
presents several contributions pushing the frontiers of GAEs and VGAEs, and showing that jointly addressing CD and LP with high accuracy is possible with these models. After reviewing key concepts in Section~\ref{s2}, we explain why GAEs and VGAEs underperform on CD in Section~\ref{s3}. We simultaneously introduce \textit{Modularity-Aware GAE and VGAE}, our solution leveraging  \textit{modularity-based clustering} concepts \cite{blondel2008louvain,brandes2007modularity,shiokawa2013fast} to improve CD while preserving the ability to identify missing edges in LP. We report an in-depth evaluation of our method in~Section~\ref{s4},~and~conclude~in~Section~\ref{s5}.



\section{Preliminaries}
\label{s2}

We consider an undirected graph $\mathcal{G} = (\mathcal{V},\mathcal{E})$ with $n$ nodes and $m$ edges. We denote its $n\times n$ adjacency matrix by $A.$ Each node $i \in \mathcal{V}$ is equipped with a feature vector $x_i \in \mathbb{R}^f$. We denote the $n \times f$ matrix having $x_i$ vectors as rows by $X.$ For a featureless graph, we set $X = I_n$,~the~identity~matrix.

\paragraph{GAE and VGAE.}
The term GAE refers to a family of unsupervised two-component models learning node embedding spaces in the absence of node labels \cite{kipf2020phd,kipf2016-2,tian2014learning,wang2016structural}. 
The first component is the \textit{encoder}, a parameterized function processing $A$ and $X$, and mapping each node $i \in \mathcal{V}$ to an \textit{embedding vector} $z_i \in \mathbb{R}^d$, with $d \ll n$.  In practice, a GNN~\cite{hamilton2020graph,kipf2016-1,zhang2018network} often acts as the encoder, i.e., $Z = \text{GNN}(A,X)$, with $Z$ the $n \times d$ matrix having $z_i$ vectors as rows.
The second component is the \textit{decoder}, estimating an adjacency matrix~$\hat{A}$ from embedding vectors: $\hat{A}=\text{Decoder}(Z)$. Decoders can be neural networks, or simpler functions, e.g., based on inner products between $z_i$ vectors~\cite{kipf2016-2,li2020graph,park2019symmetric,wang2016structural}. When training a GAE, one wishes to learn $z_i$ vectors from which reconstructing $\mathcal{G}$ should be possible. Intuitively, this would indicate that the embedding space preserves some important information about $\mathcal{G}$. 
For this purpose, model weights are trained via gradient~descent~minimization~\cite{goodfellow2016deep} of a \textit{reconstruction loss}, usually a cross entropy \cite{kipf2016-2}, evaluating the similarity~between~$\hat{A}$~and~$A.$

Introduced as probabilistic extensions of GAEs, VGAE models associate $z_i$ vectors with distributions. Notably, in the seminal VGAE from Kipf and Welling~\cite{kipf2016-2}, each vector $z_i \sim \mathcal{N}(\mu_i,\Sigma_i)$. Their model incorporates \textit{two} GNN encoders processing both $A$ and $X$: one of them learns mean vectors $\mu_i \in \mathbb{R}^d$, and the other learns variance matrices $\Sigma_i \in \mathbb{R}^{d \times d}$, for all $i \in \mathcal{V}$. Moreover, instead of a reconstruction loss, they optimize the variational \textit{evidence lower bound} (ELBO) of the model's likelihood~\cite{kingma2013vae}, using gradient ascent. Besides constituting promising generative models \cite{molecule3,molecule1,simonovsky2018graphvae}, variants of VGAEs also turned out to be effective alternatives to GAEs in some LP and CD tasks~\cite{choong2020optimizing,semiimplicit2019,kipf2016-2,salha2021fastgae,salha2020simple,salha2019-2}. 




\paragraph{Evaluation.} Over the past years, LP\footnote{\label{footnote:task} We provide more formal presentations of the LP and CD problems under consideration in Appendix A.} has become the most prominent way to evaluate the quality of embedding vectors learned from a GAE or VGAE~\cite{grover2019graphite,hao2020inductive,semiimplicit2019,huang2019rwr,pei2021generalization,pan2018arga,salha2019-2,aaai20,tran2018multi}. 
Previous work widely confirmed the effectiveness of GAEs and VGAEs~on~this~task.
Their performances are less impressive on CD\footref{footnote:task}, another important graph problem with numerous applications~\cite{cavallari2017learning,he2021community,malliaros2013clustering,sun2019vgraph,tu2018unified}.
In the presence of node embedding representations~$z_i,$ CD boils down to the common problem of clustering $n$ vectors, e.g., via a $k$-means \cite{macqueen1967some} in the embedding space.
 Nonetheless, concurring work \cite{choong2018learning,choong2020optimizing,salha2021fastgae,salha2019-1} recently pointed out the limitation of this approach for GAEs and VGAEs, and its lower performance w.r.t. simpler CD alternatives, such as the popular Louvain method learning communities by iteratively maximizing the density-based \textit{modularity} value in~the~graph~\cite{blondel2008louvain}.

While recent studies aimed to address the underwhelming performance of GAEs and VGAEs on CD, they still suffer from limitations that motivate our work.
Firstly, several studies \cite{choong2018learning,choong2020optimizing,li2020dirichlet} considered clustering-oriented probabilistic priors for VGAEs (such as Gaussian mixtures in VGAECD \cite{choong2018learning} and VGAECD-OPT \cite{choong2020optimizing}), that cannot be transposed to the deterministic GAE setting.
Secondly, a closer look at these models reveals that their empirical gains mostly stem from the addition of node features. They offer little advantage when features are absent (see Table~\ref{tab:vgaecd}). Other studies did not consider featureless graphs at all \cite{huang2019rwr,pei2021generalization,pan2018arga,park2019symmetric,aaai20}. This motivates the need to investigate GAE/VGAE-based CD on featureless graphs. Thirdly, previous studies did not try to preserve good performances on LP \cite{choong2018learning,choong2020optimizing,li2020dirichlet,wang2017mgae}. 
It is still uncertain whether one can jointly address LP and CD with accuracy in multi-task settings, which, as argued in the Introduction, is highly relevant in practice.
In conclusion, the question of improving CD with GAEs and VGAEs~remains~incompletely~addressed.

\section{Modularity-Aware GAE and VGAE for Joint LP and CD}
\label{s3}

\begin{figure}[t]
    \centering
    \includegraphics[width=0.93\textwidth]{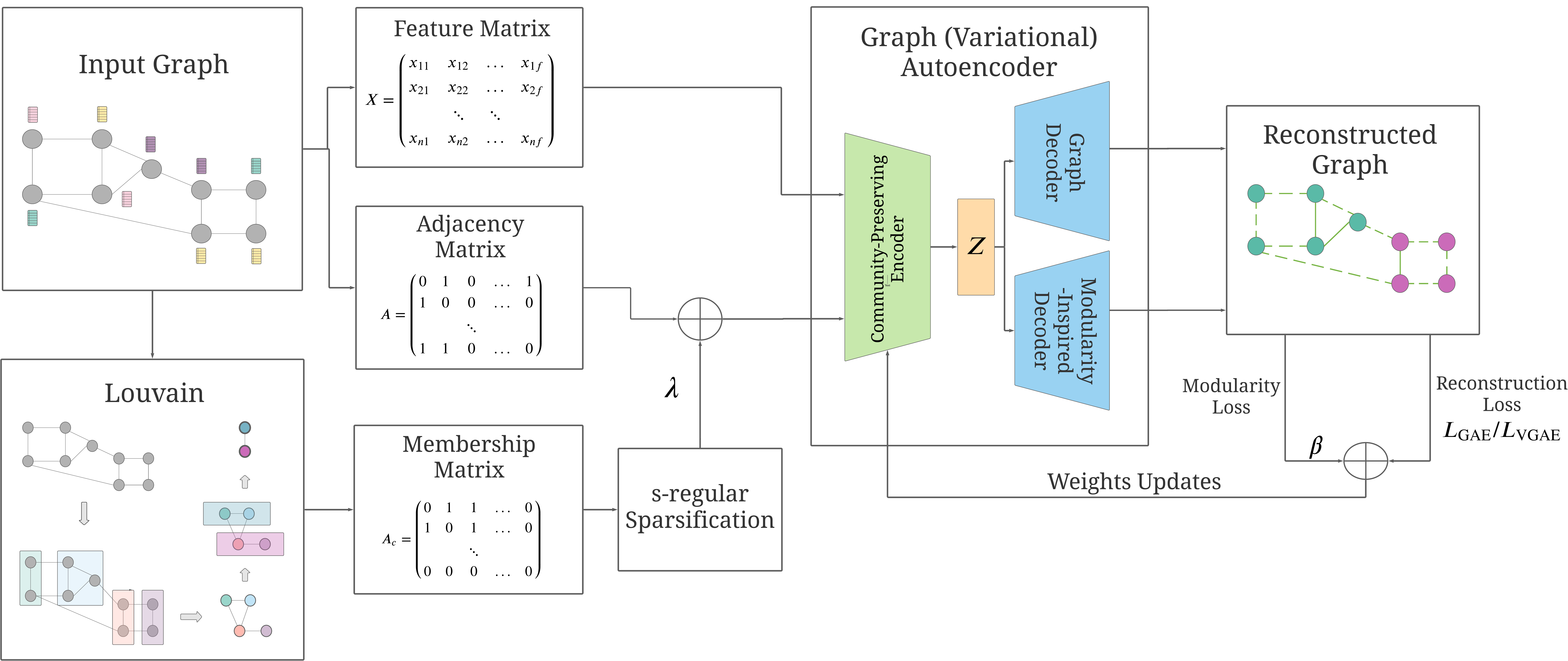}
    \caption{Overview of our proposed Modularity-Aware GAE/VGAE.
    Firstly, input graph data $A$ and $X$ are combined with the $\dregc$-regular sparsified prior community membership matrix $\Ao$, derived through iterative modularity maximization via the Louvain algorithm, as described in the first paragraph of Section~\ref{s3}. Then, they are processed by our revised community-based encoders, encoding each node $i$ as an embedding vector $z_i$ of dimension $d \ll n$. Neural weights of encoders are optimized through a procedure combining reconstruction and modularity-inspired losses, and described in the second paragraph of Section~\ref{s3}. Furthermore, other hyperparameters from this model are tuned via the method described in the third paragraph of Section~\ref{s3} and designed for joint~LP~and~CD.}
    \label{fig:magae}
\end{figure}

To address these limitations, we introduce our Modularity-Aware GAE/VGAE, illustrated~in~Figure~\ref{fig:magae}.

\paragraph{Community-Based Encoders.} Firstly, we argue that most GAEs/VGAEs leverage encoders that do not specifically aim to capture graph communities. This includes graph convolutional networks (GCN)~\cite{kipf2016-1}, which remain the most popular encoders in practice \cite{grover2019graphite,semiimplicit2019,huang2019rwr,pei2021generalization,pan2018arga,aaai20}, and encoders identifying clusters from features rather than the graph~\cite{choong2020optimizing,wang2017mgae}.
Modularity-Aware GAE and VGAE overcome this issue by incorporating~a~\textit{community-based~encoder}.

Specifically, we first obtain a partition of the node set using the Louvain method~\cite{blondel2008louvain} and store it in an $n\times n$ membership matrix $\Ac$, defined as $(\Ac)_{ij} = 1$ if nodes $i \neq j$ are in the same community, and 0 otherwise.
Then, when learning embedding vectors, we leverage this partition as a \textit{prior signal}, from which the encoder should benefit, but also have the ability to deviate. Formally, we replace~the $Z = \text{GNN}(A,X)$ component\footnote{For clarity of exposition we discuss the deterministic GAE framework. Our modifications equally apply to the VGAE framework, for which $Z$ has to be replaced by Gaussian parameters (see Section~\ref{s2}).} by: $Z = \text{GNN}(A + \lambda A_s,X)$,  where $\lambda \in \mathbb{R}^+$ and $s \in \mathbb{N}^+$ are hyperparameters, and where $A_s$ is a $\dregc$-regular sparsified\footnote{In $A_s$, nodes are only connected to $s$ fixed and randomly selected neighbors from their community. This sparsification permits speeding up GNN message passing operations in practice~\cite{kipf2016-1}.} version of $\Ac.$
This change alters the GNN~\textit{message~passing~scheme}. Nodes will now aggregate information from their neighbors \textit{and} some nodes of their prior community ($\lambda$ balances the importance of these two information sources). Therefore, nodes from the same prior community will tend to have more similar embedding vectors than with~a~standard~GAE~or~VGAE.

Besides its simplicity and good performance on CD~\cite{salha2021fastgae}, our justification for using Louvain as a prior is threefold. Firstly, it automatically selects the relevant number of prior communities to consider. Secondly, it runs in $O(n\log n)$ time \cite{blondel2008louvain} and, therefore,
scales to graphs with millions of nodes. Thirdly, it optimizes a \textit{modularity} criterion that complements
the encoding-decoding paradigm. We will show in Section~\ref{s4} that learning representations
from complementary criteria is beneficial. Nonetheless, our framework remains valid for any alternative method~providing~prior~communities.

\paragraph{Modularity-Inspired Losses.} Previous models were also \textit{trained} in a fashion that, by design, favors LP over CD. The cross entropy and ELBO losses involve the reconstruction of \textit{node pairs} from the embedding space \cite{kipf2016-2}. However, a good reconstruction of \textit{local} pairwise connections does not necessarily imply a good reconstruction of the \textit{global} community structure~\cite{liu2019much,wang2017community}. Consequently, in Modularity-Aware GAE (respectively, VGAE), we minimize (resp., maximize), using gradient descent (resp., gradient ascent), an alternative function that subtracts (resp., adds) the following \textit{global regularizer} to the cross entropy (resp., ELBO) term:
$\frac{\beta}{2m} \sum_{i,j=1}^n [A_{ij} - \frac{d_id_j}{2m}]  e^{-\gamma \Vert z_i - z_j \Vert^2_2},$ with $d_i$ the degree \cite{hamilton2020graph} of node $i \in \mathcal{V}$  and two hyperparameters $\beta \in \mathbb{R}^+$ and $\gamma \in \mathbb{R}^+.$

A soft and differentiable version of the \textit{modularity}~\cite{Newman8577} (independent of any ground truth community), this regularizer aims to push closer vectors $z_i$ of densely connected parts of the graph, and, therefore, to permit a $k$-means-based detection of communities with higher density. Several studies out of the GAE/VGAE scope emphasized the effectiveness of comparable approaches for learning community-preserving representations~\cite{lobov2019unsupervised,wang2017community,yang2016modularity}. On the other hand, the remaining presence of the local cross entropy (resp., ELBO) in our optimized loss aims to preserve good performances on LP. $\beta$~balances the relative importance of the global regularizer. $\gamma$ regulates the magnitude of $\Vert z_i - z_j \Vert^2_2$ in the exponential term, which tends to 1 when $z_i$ and $z_j$ get closer, and to 0 when they move apart. 

\paragraph{Hyperparameter Selection.} GAEs and VGAEs include several important hyperparameters such as dropout and learning rates \cite{kipf2016-2} (our models also introduce $\lambda$, $s$, $\beta$,~and~$\gamma$). In previous studies, their selection procedure was sometimes solely based on LP validation~sets~\cite{salha2021fastgae,salha2019-1}. However, optimal values for CD might differ from those for LP, partly explaining the low~performance~on~CD. In this paper, we consider an alternative hyperparameter selection procedure. As detailed in Appendix~A, the hyperparameters selected for our models are chosen by maximizing the average of: (1) an \textit{Area under the ROC Curve (AUC) score} computed on an LP validation set, and (2)~the \textit{modularity} score computed from the communities extracted from final vectors $z_i$, via a $k$-means.  We expect this dual criterion to identify hyperparameters that will be jointly relevant for LP and CD in a~multi-task~setting. 

\section{Experimental Evaluation}
\label{s4}

We now report results from an in-depth experimental evaluation of our method. Our code is available on GitHub: \texttt{\href{https://github.com/GuillaumeSalhaGalvan/modularity_aware_gae}{https://github.com/GuillaumeSalhaGalvan/modularity\_aware\_gae}}.

\paragraph{Setting.} For evaluation, we consider a ``pure'' CD problem, as well as a multi-task LP/CD problem, on seven graphs of various origins and sizes (from 1124 to 2.5~million~nodes). For both problems and all graphs, we compare our approach to 12 baselines, including the Louvain method~\cite{blondel2008louvain}, standard GAE/VGAE models~\cite{kipf2016-2} with varying encoders, and existing extensions of GAEs/VGAEs for CD. For brevity, we report technical details on tasks, datasets, models, and hyperparameters in~Appendix~A.

\paragraph{Results on CD.}
CD results from Table~\ref{tab:allresults} confirm the discussed limitations of standard GAE/VGAE, which Louvain outperforms on 5 of 7 featureless graphs (e.g., +7.45 Adjusted Mutual Information (AMI) points for Louvain on Pubmed).
On the contrary, our Modularity-Aware GAE/VGAE almost always surpass the Louvain method, \textit{and} the use of a standard GAE/VGAE (e.g., with a top 21.64\% AMI on the largest Album graph).
Interestingly, combining Louvain and a GAE/VGAE into our Modularity-Aware models is beneficial even when the GAE/VGAE initially outperforms Louvain (e.g., for Cora-Large). This confirms that modularity-based clustering \textit{à la} Louvain complements the encoding-decoding paradigm, and that leveraging complementary criteria is empirically beneficial.
We also compare favorably to other baselines in most experiments (e.g., +2.11 AMI points w.r.t. VGAECD-OPT~\cite{choong2020optimizing} on Cora with features), with or without the addition~of~node~features. Figure~\ref{visucora} provides a visualization of node embedding representations learned by our models.

\paragraph{Results on Multi-Task CD/LP.} We now assess whether improving CD implies deteriorating the effectiveness on LP. The last columns of Table~\ref{tab:allresults} confirm the ability of Modularity-Aware GAE/VGAE to preserve good performances on LP (we achieve comparable scores w.r.t standard GAE/VGAE on all graphs).
While performances on CD decrease slightly w.r.t. pure CD (an expected result, as some edges are masked during training for the purpose of LP), we continue to outperform baselines in most experiments. This demonstrates the effectiveness of our approach at jointly addressing~CD~and~LP. 

\paragraph{Discussion on Model Components.} For most models, using a linear encoder~\cite{salha2020simple} gives competitive LP/CD results w.r.t. a 2-layer GCN~\cite{kipf2016-2}. Also, VGAE models often outperform their GAE counterparts, even though scores are relatively close.
Our proposed hyperparameter selection procedure had a noticeable impact on the choices of $\lambda$, $\beta$, $\gamma$, and $\dregc$, as well as on the required number of training iterations, which we illustrate in Figure~\ref{fig:optimization}. In such cases, optimal values for joint LP and CD differ from those for LP only. Lastly, one might wonder whether our performance gains mainly come from our novel encoder or our regularized loss. Figure~\ref{fig:ablation} reports an \textit{ablation study}, consisting in training variant versions of Modularity-Aware VGAEs with one component only (i.e., the novel encoder but not the regularized loss, or vice versa). We show that incorporating any of these two individual contributions into the VGAE improves CD, and that their simultaneous use~leads~to~the~best~results.


\section{Conclusion}
\label{s5}

In this paper, we introduced a well-performing approach for joint CD and LP with GAEs and VGAEs. We demonstrated its effectiveness through in-depth experimental validation. Our work paves the way for various future research, including replacing Louvain with other prior methods, using our regularizer in conjunction with other reconstruction losses (e.g., ELBO variants computed from Gaussian mixtures~\cite{choong2018learning,choong2020optimizing}), and extending our approach to dynamic graphs. 
The journal version\footref{footnote:journal} of this work \cite{salhagalvan2022modularity} includes several additional extensions as well as results, omitted here for brevity. This includes further comparisons to non-GAE/VGAE methods, a spectral analysis of our message passing operator, and discussions on how this research helps the music streaming service Deezer address real-world multi-task LP and CD problems for~music~recommendation~purposes.

\bibliographystyle{ACM-Reference-Format}
\bibliography{references}

\newpage

\appendix
\section*{Appendix}

This appendix provides details on our experimental setting in Appendix A, and complementary tables and figures from our experiments in Appendix B.

\section{Experimental Setting}

\paragraph{Datasets.} We consider seven graphs of various origins, characteristics, and sizes. Firstly, we study the \textit{Cora} ($n = 2708$, $m = 5429$), \textit{Citeseer} ($n = 3327$, $m = 4732$) and \textit{Pubmed} ($n = 19717$, $m = 44338$) citation networks \cite{kipf2016-1}, \textit{with} and \textit{without} node features that correspond to bag-of-words vectors of dimensions $f = 1433$, $3703$, and $500$, respectively. In these datasets, nodes are clustered in 6, 7, and 3 topic classes, respectively, acting as the communities to be detected. These graphs are commonly used to evaluate GAEs and VGAEs. We, therefore, see value in studying them as well, especially in their \textit{featureless} version where previous GAE and VGAE extensions fall short on CD.

In addition, we consider a larger version of Cora, referred to as \textit{Cora-Large} ($n = 23166$, $m = 91500$)~\cite{salha2020simple}. Nodes are documents clustered in 70 topic-related communities. Additionally, we consider the \textit{Blogs web graph} ($n = 1224$, $m = 19025$)~\cite{salha2020simple}. Nodes correspond to webpages of political blogs connected through hyperlinks, and clustered in two communities corresponding to politically left-leaning or right-leaning blogs.
Thirdly, we examine the \textit{SBM} graph ($n = 100000$, $m = 1498844$), generated from a \textit{stochastic block model}, i.e., a generative model for community-based random graphs~\cite{abbe2017community}. Nodes are clustered in 100 ground truth communities of 1000 nodes each. Nodes from the same community are connected with probability $p = 2 \times 10^{-2}$, while nodes from different
communities are connected with probability $q = 2 \times 10^{-4} < p$. Albeit being synthetic, this graph includes actual node communities by design, and is, therefore, relevant to evaluate CD~methods.

Lastly, we consider \textit{Album} ($n = 2503985$, $m = 25039155$) a private graph provided by the music streaming service Deezer. Nodes are \textit{music albums} available on this service, connected through an undirected edge when they are regularly \textit{co-listened} to by users. The service is jointly interested in (1)~predicting new connections in the graph, corresponding to new albums pairs that users would enjoy listening to together; and (2)~learning groups of similar albums, with the aim of providing usage-based recommendations (i.e., if users listen to several albums from a community, other unlistened albums from this same community could be recommended to them). In such an application, learning album representations that would \textit{jointly} enable effective LP and CD would therefore be desirable. For evaluation, communities will be compared to a ground truth clustering of albums in 20 groups defined by their main \textit{music genre}, allowing us to assess the musical homogeneity of node~communities.

\paragraph{Tasks.} For each of these graphs, we assess the performance of our models on two~downstream~tasks.
\begin{itemize}
    \item \textbf{Task 1:} We first examine a ``pure'' \textit{CD} task, consisting in the extraction of a partition of the node set $\mathcal{V}$ which ideally agrees with the ground truth communities of each graph. Communities will be retrieved by running a $k$-means (with $k$-means++ initialization~\cite{arthur2017kmeans}) in the final embedding space of each model to cluster the vectors $z_i$, with $k$ matching the known number of communities; except for some baseline methods that explicitly incorporate another strategy to partition nodes. We compare the obtained partitions to the ground truth using the \textit{Adjusted Mutual Information
(AMI)} and \textit{Adjusted Rand Index (ARI)} scores\footnote{\label{footnote:sklearn} Scores are computed via scikit-learn, using formulas provided in the sklearn.metrics documentation \cite{pedregosa2011scikit}.}. 
\item \textbf{Task 2:} We also study a \textit{joint LP and CD} task. In such a \textit{multi-task} setting, we learn all node embedding spaces from \textit{incomplete} versions of the seven graphs, where 15\% of edges were randomly masked. We create a validation and
a test set from these masked edges (from 5\% and 10\% of edges, respectively) and the same number of randomly picked unconnected node pairs acting as ``non-edge'' negative pairs. Then, using decoder predictions $\hat{A}_{
ij}$ computed from vectors $z_i$ and $z_j,$ we evaluate each model's ability to distinguish edges from non-edges, i.e., LP, from the embedding space, using the \textit{Area
Under the ROC Curve (AUC)} and \textit{Average Precision (AP)} scores\footref{footnote:sklearn}. Jointly, we evaluate the CD performance obtained from such incomplete graphs, using the same methodology and scores as in Task~1.
\end{itemize}

In the case of Task~2, we expect AMI and ARI scores to decrease w.r.t. Task~1, as models will only observe \textit{incomplete} versions of the graphs when learning embedding spaces. With Task~2, we aim to assess whether improving CD inevitably leads to deteriorating performances~on~LP.

\paragraph{Models: Details on the Hyperparameter Selection Procedure.} For these two tasks and seven graphs, we compare the performances of our proposed Modularity-Aware GAE and VGAE to standard GAE and VGAE and to several other baselines. All models described below will verify $d = 16$ (the journal version of this work also discusses results obtained with $d \in \{32, 64\}$, which lead to similar conclusions as $d=16$). 
We choose other hyperparameters using the \textit{selection procedure} mentioned in Section~\ref{s3}, and further described in the next paragraph.

Foremost, as CD is an unsupervised task, we cannot rely on train/validation/test splits as for the supervised LP classification task\footnote{Ground truth communities are \textit{unavailable} during training. They will only be revealed for model evaluation, to compare the agreement of the node partition inferred by each model to the ground truth partition.}. Consistently with our other contributions, we rather rely on the \textit{modularity}~\cite{Newman8577}, an unsupervised density-based criterion computed independently of ground truth communities. Precisely, we select hyperparameters that~maximize~the~average~of:
        \begin{itemize}
            \item the AUC obtained for LP on the validation set of Task 2; 
            \item the modularity:
$Q = \frac{1}{2m} \sum_{i,j=1}^n [A_{ij} - \frac{d_id_j}{2m}] \delta(i,j)$, computed from the communities extracted by running a $k$-means on the final vectors $z_i,$ learned from the train graph of Task 2. In this equation, $\delta(i,j) = 1$ if nodes $i$ and $j$ belong to the same community~and~$0$~otherwise.
        \end{itemize}
We expect this dual criterion to identify
hyperparameters jointly relevant to LP and CD.

\begin{table}[t]
\centering
\caption{Complete list of optimal hyperparameters of Modularity-Aware GAE and VGAE models.}
    \label{tab:hyperparameterstable}
   \resizebox{0.95\textwidth}{!}{
\begin{tabular}{c|cccccccc}
\toprule
\textbf{Dataset} & \textbf{Learning} & \textbf{Number of} & \textbf{Dropout} & \textbf{Use of FastGAE \cite{salha2021fastgae}} & $\lambda$ & $\beta$ & $\gamma$ & $\dregc$ \\
& \textbf{rate} & \textbf{iterations} & \textbf{rate} & \textbf{(if yes: subgraphs size)} & & & \\
\midrule
\midrule
\textbf{Blogs} & 0.01 & 200 & 0.0 & No & 0.5 & 0.75 & 2 & 10\\
\textbf{Cora (featureless)}  & 0.01 & 500 & 0.0 & No & 0.25 & 1.0 & 0.25 & 1\\
\textbf{Cora (with features)}  & 0.01 & 300 & 0.0 & No & 0.001 & 0.01 & 1 & 1\\
\textbf{Citeseer (featureless)}  & 0.01 & 500 & 0.0 & No & 0.75 & 0.5 & 0.5 & 2\\
\textbf{Citeseer (with features)}  & 0.01 & 500 & 0.0 & No & 0.75 & 0.5 & 0.5 & 2\\
\textbf{Pubmed (featureless)}  & 0.01 & 500 & 0.0 & No & 0.1 & 0.5 & 0.1 & 5\\ 
\textbf{Pubmed (with features)}  & 0.01 & 700 & 0.0 & No & 0.1 & 0.5 & 10 & 2\\
\textbf{Cora-Large}  & 0.01 & 500 & 0.0 & No & 0.001 & 0.1 & 0.1 & 10 \\
\textbf{SBM}  & 0.01 & 300 & 0.0 & Yes (10 000) & 0.5 & 0.1 & 2 & 10 \\
\textbf{Album} & 0.005 & 600 & 0.0 & Yes (10 000) & 0.25 & 0.25 & 1 & 5\\
\bottomrule
\end{tabular}}
\end{table}

\paragraph{Models: Modularity-Aware GAE and VGAE.} We trained two versions of our Modularity-Aware GAE and VGAE: one with the \textit{linear encoder} proposed by Salha et al.~\cite{salha2020simple}, and one with the \textit{2-layer GCN encoder} used by Kipf and Welling~\cite{kipf2016-2}. The latter encoder includes a 32-dimensional hidden layer. As most GAE/VGAE models, we use a simple inner product decoder: $\hat{A}_{ij} = \sigma(z^T_i z_j)$. 

During training, we used the Adam optimizer~\cite{kingma2014adam}, without dropout (but we tested models with dropout values in $\{0,0.1,0.2\}$ in our grid search optimization). For each graph, we considered learning rates from the grid $\{0.001,0.005,0.01,0.05,0.1,0.2\}$, number of training iterations in $\{100, 200, 300, ..., 800\}$, with $\lambda \in \{0, 0.01,0.05, 0.1, 0.2, 0.3, ..., 1.0\}$,  $\beta \in \{0, 0.01,0.05, 0.1, 0.25, 0.5, 1.0, 1.5, 2.0\}$, $\gamma \in \{0.1, 0.2, 0.5, 1.0, 2, 5, 10\}$ and $\dregc \in \{1, 2, 5, 10\}$. The best hyperparameters for each graph are reported in Table~\ref{tab:hyperparameterstable}. We adopted the same optimal hyperparameters for GAE \textit{and} VGAE variants. Lastly, as the exact loss computation was computationally infeasible for our two largest graphs, SBM and Album, their corresponding models were trained by using the FastGAE method \cite{salha2021fastgae}, approximating losses by reconstructing degree-based sampled subgraphs of $n = 10000$ nodes (a different one at each training iteration).

We used Tensorflow~\cite{abadi2016tensorflow}, training our models (as well as GAE/VGAE baselines described below) on an NVIDIA GTX 1080 GPU, and running other operations on a double Intel Xeon Gold 6134 CPU\footnote{On our machines, running times of the Modularity-Aware GAE and VGAE were comparable to running times of their standard GAE and VGAE counterparts. For example, training each variant of VGAE on the Pubmed graph for 500 training iterations and with $\dregc = 5$ takes 25 minutes on a single GPU (without FastGAE).}.

\paragraph{Models: Standard GAE and VGAE.} We examine two variants of the standard GAE and VGAE: one with 2-layer GCN encoders with a 32-dimensional hidden layer (which is equal to the GAE and VGAE from Kipf and Welling~\cite{kipf2016-2}) and one with a linear encoder (which is equal to the linear GAE and VGAE from Salha et al.~\cite{salha2020simple}). We note that these models are particular cases of our Modularity-Aware GAE/VGAE with GCN or linear encoder and with $\lambda = 0$ and $\beta = 0$. As for our Modularity-Aware models, LP is performed from inner product decoding, and CD via a $k$-means on vectors $z_i$. We selected similar learning rates and numbers of iterations to the values~reported~in~Table~\ref{tab:hyperparameterstable}.

\paragraph{Models: Other Baselines.}
We also report experiments on VGAECD \cite{choong2018learning}, a \textit{VGAE for CD} model that replaces Gaussian priors by learnable \textit{Gaussian mixtures}. Such a change permits recovering communities from node embedding spaces without relying on an additional $k$-means step. We also tested VGAECD-OPT,  an improved version of VGAECD by the same authors~\cite{choong2020optimizing}. Specifically, VGAECD-OPT replaces GCN encoders with linear models. It also adopts a different optimization procedure based on neural expectation-maximization, which guarantees that communities do not collapse during training and experimentally leads to better performances~\cite{choong2020optimizing}. We set similar hyperparameters to the above other GAE/VGAE-based models. In all models, the number of Gaussian mixtures matches the ground truth number of communities in each graph.

Besides, we also report experiments on the \textit{Dirichlet Graph Variational Autoencoder} (DGVAE)~\cite{li2020dirichlet}, another extension of VGAE which uses Dirichlet distributions as priors on latent vectors, acting as indicators of community membership. We set similar learning rates and layer dimensions to the above GAE/VGAE-based models. In the case of DGVAE, we use 2-layer GCN encoders for consistency with other models in our experiments. We nonetheless acknowledge that the authors also proposed another encoder, denoted Heatts in their paper (but unavailable in their public code at the time of writing) that could replace GCNs both in DGVAE and in Modularity-Aware GAE and VGAE. 

We also examine the \textit{Adversartially-Regularized (Variational) Graph Autoencoder} (ARGA and ARVGA) models \cite{pan2018arga}, that incorporate an adversarial regularization scheme to GAE and VGAE, with similar hyperparameters as previous models. ARGA and ARVGA emerged as some of the most cited GAE/VGAE extensions and, while they were not specifically introduced for CD, Pan et al.~\cite{pan2018arga} reported empirical gains on this task w.r.t.~standard~GAE/VGAE, on graphs with node features.

For completeness, we add three baselines not utilizing the autoencoder paradigm. We report results obtained from the popular node embedding methods \textit{node2vec}~\cite{grover2016node2vec} and \textit{DeepWalk} \cite{perozzi2014deepwalk}, training models from 10 random walks with length 80 per node, a window size of 5 and on a single epoch. For node2vec, we further set $p=q=1$. We use a similar strategy as GAEs/VGAEs ($k$-means/inner products) for CD and LP from embedding spaces. Lastly, we also compare to the \textit{Louvain} method~\cite{blondel2008louvain} for CD. 
We see value in comparing to a direct use of Louvain, as this method is directly leveraged in our Modularity-Aware GAE/VGAE as a pre-processing step for the computation~of~$A_c$~and~$\Ao$.  

\newpage

\section{Figures and Tables}

We now provide complementary tables and figures from our experiments. Table~\ref{tab:coraresults} details complete results for the Cora dataset and Table~\ref{tab:allresults} reports more summarized results for several more graphs.

\begin{table}[h]
\begin{center}
\begin{small}
\centering
\caption{Results for Task 1 and Task 2 on the featureless Cora graph, using Modularity-Aware GAE/VGAE with Linear and GCN encoders, their standard GAE/VGAE counterparts, and other baselines. All node embedding models learn vectors of dimension $d =16$. Scores are averaged over 100 runs. LP results are reported from test sets. \textbf{Bold} numbers correspond to the best performance for each score. Scores \textit{in italic} are within one standard deviation range from~the~best~score.}
    \label{tab:coraresults}
   \resizebox{0.99\textwidth}{!}{
\begin{tabular}{r||cc||cc|cc}
\toprule
\textbf{Models} &  \multicolumn{2}{c}{\textbf{Task 1: Community Detection}} & \multicolumn{4}{c}{\textbf{Task 2: Joint Link Prediction and Community Detection}}\\
(Dimension $d=16$) & \multicolumn{2}{c}{\textbf{on complete graph}} & \multicolumn{4}{c}{\textbf{on graph with 15\% of edges being masked}}\\
\midrule
 & \textbf{AMI (in \%)} & \textbf{ARI (in \%)} &  \textbf{AMI (in \%)} &  \textbf{ARI (in \%)} &  \textbf{AUC (in \%)} & \textbf{AP (in \%)} \\ 
\midrule
\midrule
\underline{\textit{Modularity-Aware GAE/VGAE Models}} &  &  &  &  &  &  \\
Linear Modularity-Aware VGAE & \textbf{46.65} $\pm$ \textbf{0.94} & \textit{39.43} $\pm$ \textit{1.15} & \textit{42.86} $\pm$ \textit{1.65} & \textit{34.53} $\pm$ \textit{1.97} & \textit{85.96} $\pm$ \textit{1.24} & \textit{87.21} $\pm$ \textit{1.39} \\
Linear Modularity-Aware GAE & \textit{46.58} $\pm$ \textit{0.40} & \textbf{39.71} $\pm$ \textbf{0.41} & \textbf{43.48} $\pm$ \textbf{1.12} & \textbf{35.51} $\pm$ \textbf{1.20} & \textbf{87.18} $\pm$ \textbf{1.05} & \textit{88.53} $\pm$ \textit{1.33} \\
GCN-based Modularity-Aware VGAE & 43.25 $\pm$ 1.62 & 35.08 $\pm$ 1.88 & 41.03 $\pm$ 1.55 & \textit{33.43} $\pm$ \textit{2.17} & 84.87 $\pm$ 1.14 & 85.16 $\pm$ 1.23  \\
GCN-based Modularity-Aware GAE & 44.39 $\pm$ 0.85 & 38.70 $\pm$ 0.94 & 41.13 $\pm$ 1.35 & \textit{35.01} $\pm$ \textit{1.58} & \textit{86.90} $\pm$ \textit{1.16} & \textit{87.55} $\pm$ \textit{1.26} \\
\midrule
\underline{\textit{Standard GAE/VGAE Models}} &  &  &  &  &  &  \\
Linear VGAE & 37.12 $\pm$ 1.46 & 26.83 $\pm$ 1.68 & 32.22 $\pm$ 1.76 & 21.82 $\pm$ 1.80 & 85.69 $\pm$ 1.17 & \textbf{89.12} $\pm$ \textbf{0.82} \\
Linear GAE & 35.05 $\pm$ 2.55 & 24.32 $\pm$ 2.99 & 28.41 $\pm$ 1.68 & 19.45 $\pm$ 1.75 & 84.46 $\pm$ 1.64 & \textit{88.42} $\pm$ \textit{1.07} \\
GCN-based VGAE & 34.36 $\pm$ 3.66 & 23.98 $\pm$ 5.01 & 28.62 $\pm$ 2.76 & 19.70 $\pm$ 3.71 & 85.47 $\pm$ 1.18 & \textit{88.90} $\pm$ \textit{1.11} \\
GCN-based GAE & 35.64 $\pm$ 3.67 & 25.33 $\pm$ 4.06 & 31.30 $\pm$ 2.07 & 19.89 $\pm$ 3.07 & 85.31 $\pm$ 1.35 & \textit{88.67} $\pm$ \textit{1.24} \\
\midrule
\midrule
\underline{\textit{Other Baselines}} &  &  &  &  &  &  \\
Louvain & 42.70 $\pm$ 0.65 & 24.01 $\pm$ 1.70 & 39.09 $\pm$ 0.73 & 20.19 $\pm$ 1.73 & -- & -- \\
VGAECD& 36.11 $\pm$ 1.07 & 27.15 $\pm$ 2.05 & 33.54 $\pm$ 1.46 & 24.32 $\pm$ 2.25 & 83.12 $\pm$ 1.11 & 84.68 $\pm$ 0.98 \\
VGAECD-OPT & 38.93 $\pm$ 1.21 & 27.61 $\pm$ 1.82 & 34.41 $\pm$ 1.62 & 24.66 $\pm$ 1.98 & 82.89 $\pm$ 1.20 & 83.70 $\pm$ 1.16 \\
ARGVA & 34.97 $\pm$ 3.01 & 23.29 $\pm$ 3.21 & 28.96 $\pm$ 2.64 & 19.74 $\pm$ 3.02 & 85.85 $\pm$ 0.87 & \textit{88.94} $\pm$ \textit{0.72} \\
ARGA & 35.91 $\pm$ 3.11 & 25.88 $\pm$ 2.89 & 31.61 $\pm$ 2.05 & 20.18 $\pm$ 2.92 & 85.95 $\pm$ 0.85 & \textit{89.07} $\pm$ \textit{0.70} \\
DVGAE & 35.02 $\pm$ 2.73 & 25.03 $\pm$ 4.32 & 30.46 $\pm$ 4.12 & 21.06 $\pm$ 5.06 & 85.58 $\pm$ 1.31 & \textit{88.77} $\pm$ \textit{1.29} \\
DeepWalk & 36.58 $\pm$ 1.69 & 27.92 $\pm$ 2.93 & 30.26 $\pm$ 2.32 & 20.24 $\pm$ 3.91 & 80.67 $\pm$ 1.50 & 80.48 $\pm$ 1.28 \\
node2vec & 41.64 $\pm$ 1.25 & 34.30 $\pm$ 1.92 & 36.25 $\pm$ 1.38 & 29.43 $\pm$ 2.21 & 82.43 $\pm$ 1.23 & 81.60 $\pm$ 0.91 \\
\bottomrule
\end{tabular}}
\end{small}
\end{center}
\end{table}

\begin{table}[h]
\begin{center}
\begin{small}
\centering
\caption{Summarized results for Task 1 and Task 2 on all graphs. For each graph, for brevity, we only report the \textbf{best} Modularity-Inspired model (best on Task 2, among GCN \textbf{or} Linear encoder, and GAE \textbf{or} VGAE), its standard counterpart, and a comparison to the Louvain baseline as well as the best other baseline (among VGAECD, VGAECD-OPT, ARGA, ARGVA, DVGAE, DeepWalk, and node2vec). All node embedding models learn vectors of dimension $d =16$. Scores are averaged over 100 runs except for SBM and Album (10 runs). \textbf{Bold} numbers correspond to the best performance for each score. Scores \textit{in italic} are within one standard deviation range from the best score.}
    \label{tab:allresults}
   \resizebox{1.0\textwidth}{!}{
\begin{tabular}{c|r||cc||cc|cc}
\toprule
\textbf{Datasets} & \textbf{Models} &  \multicolumn{2}{c}{\textbf{Task 1: Community Detection}} & \multicolumn{4}{c}{\textbf{Task 2: Joint Link Prediction and Community Detection}}\\
& (Dimension $d=16$) & \multicolumn{2}{c}{\textbf{on complete graph}} & \multicolumn{4}{c}{\textbf{on graph with 15\% of edges being masked}}\\
\midrule
&  & \textbf{AMI (in \%)} & \textbf{ARI (in \%)} &  \textbf{AMI (in \%)} &  \textbf{ARI (in \%)} &  \textbf{AUC (in \%)} & \textbf{AP (in \%)} \\ 
\midrule
\midrule
& GCN-based Modularity-Aware VGAE & \textbf{73.74} $\pm$ \textbf{1.32} & \textbf{82.78} $\pm$ \textbf{1.27} & \textbf{70.42} $\pm$ \textbf{1.28} & \textbf{79.80} $\pm$ \textbf{1.12} & \textbf{91.67} $\pm$ \textbf{0.39} & \textit{92.37} $\pm$ \textit{0.41} \\
& GCN-based Standard VGAE & \textit{73.42} $\pm$ \textit{0.95} & \textit{82.58} $\pm$ \textit{0.93} & 66.90 $\pm$ 3.32 & \textit{77.23} $\pm$ \textit{3.89} & \textit{91.64} $\pm$ \textit{0.42} & \textbf{92.52} $\pm$ \textbf{0.51} \\ 
\textbf{Blogs} & Louvain & 63.43 $\pm$ 0.86 & 76.66 $\pm$ 0.70 & 57.25 $\pm$ 1.67 & 73.00 $\pm$ 1.56 & -- & -- \\
& \underline{Best other baseline:} &  &  &  &  &  & \\
& node2vec & 72.88 $\pm$ 0.87 & 82.08 $\pm$ 0.73 & 67.64 $\pm$ 1.23 & 77.03 $\pm$ 1.85 & 83.63 $\pm$ 0.34 & 79.60 $\pm$ 0.61\\
\midrule
& Linear Modularity-Aware GAE & \textbf{46.58} $\pm$ \textbf{0.40} & \textbf{39.71} $\pm$ \textbf{0.41} & \textbf{43.48} $\pm$ \textbf{1.12} & \textbf{35.51} $\pm$ \textbf{1.20} & \textbf{87.18} $\pm$ \textbf{1.05} & \textbf{88.53} $\pm$ \textbf{1.33} \\
& Linear Standard GAE & 35.05 $\pm$ 2.55 & 24.32 $\pm$ 2.99 & 28.41 $\pm$ 1.68 & 19.45 $\pm$ 1.75 & 84.46 $\pm$ 1.64 & \textit{88.42} $\pm$ \textit{1.07} \\
\textbf{Cora} & Louvain & 42.70 $\pm$ 0.65 & 24.01 $\pm$ 1.70 & 39.09 $\pm$ 0.73 & 20.19 $\pm$ 1.73 & -- & -- \\
& \underline{Best other baseline:} &  & &  &  & &  \\
& node2vec & 41.64 $\pm$ 1.25 & 34.30 $\pm$ 1.92 & 36.25 $\pm$ 1.38 & 29.43 $\pm$ 2.21 & 82.43 $\pm$ 1.23 & 81.60 $\pm$ 0.91 \\
\midrule
& Linear Modularity-Aware VGAE & \textbf{52.43} $\pm$ \textbf{1.87} & \textbf{44.82} $\pm$ \textbf{3.12} & \textbf{49.48} $\pm$ \textbf{2.15} & \textbf{43.05} $\pm$ \textbf{3.51} & \textbf{93.10} $\pm$ \textbf{0.88} & \textbf{94.06} $\pm$ \textbf{0.75} \\
\textbf{Cora} & Linear Standard VGAE & 49.98 $\pm$ 2.40 & \textit{43.15} $\pm$ \textit{4.35} & 46.90 $\pm$ 1.43 & 38.24 $\pm$ 3.56 & \textit{93.04} $\pm$ \textit{0.80} & \textit{94.04} $\pm$ \textit{0.75} \\ 
\textbf{with} & Louvain & 42.70 $\pm$ 0.65 & 24.01 $\pm$ 1.70 & 39.09 $\pm$ 0.73 & 20.19 $\pm$ 1.73 & -- & -- \\
\textbf{features} & \underline{Best other baseline:} &  &  &  &  &  & \\
& VGAECD-OPT & 50.32 $\pm$ 1.95 & \textit{43.54} $\pm$ \textit{3.23} & 47.83 $\pm$ 1.64 & 39.45 $\pm$ 3.53 & \textit{92.25} $\pm$ \textit{1.07} & 92.60 $\pm$ 0.91 \\
\midrule

& Linear Modularity-Aware VGAE & 21.28 $\pm$ 1.03 & \textbf{15.39} $\pm$ \textbf{1.06} & 19.05 $\pm$ 1.47 & \textbf{12.19} $\pm$ \textbf{1.38} & \textbf{80.84} $\pm$ \textbf{1.64} & \textbf{84.21} $\pm$ \textbf{1.21} \\
& Linear Standard VGAE & 13.83 $\pm$ 1.00 & 8.31 $\pm$ 0.89 & 11.11 $\pm$ 1.10 & 5.87 $\pm$ 0.87 & 78.26 $\pm$ 1.55 & \textit{82.93} $\pm$ \textit{1.39} \\ 
\textbf{Citeseer} & Louvain & \textbf{24.72} $\pm$ \textbf{0.27} & 9.21 $\pm$ 0.75 & \textbf{22.71} $\pm$ \textbf{0.47} & 7.70 $\pm$ 0.67 & -- & -- \\
& \underline{Best other baseline:} &  &  &  &  &  & \\
& node2vec & 18.68 $\pm$ 1.13  & \textit{14.93} $\pm$ \textit{1.15} & 14.40 $\pm$ 1.18 & \textit{12.13} $\pm$ \textit{1.53} & 76.05 $\pm$ 2.12 & 79.46 $\pm$ 1.65 \\
\midrule

& Linear Modularity-Aware VGAE  & \textbf{25.11} $\pm$ \textbf{0.94} & \textbf{15.55} $\pm$ \textbf{0.60} & \textit{22.21} $\pm$ \textit{1.24} & \textbf{12.59} $\pm$ \textbf{1.25} & 86.54 $\pm$ 1.20 & 88.07 $\pm$ 1.22 \\
\textbf{Citeseer} & Linear Standard VGAE & 17.80 $\pm$ 1.61 & 6.01 $\pm$ 1.46 & 17.38 $\pm$ 1.43 & 6.10 $\pm$ 1.51 & \textbf{89.08} $\pm$ \textbf{1.19} & \textbf{91.19} $\pm$ \textbf{0.98} \\ 
\textbf{with} & Louvain & 24.72 $\pm$ 0.27 & 9.21 $\pm$ 0.75 & \textbf{22.71} $\pm$ \textbf{0.47} & 7.70 $\pm$ 0.67 & -- & -- \\
\textbf{features} & \underline{Best other baseline:}&  &  &  &  &  & \\
& DVGAE & 20.09 $\pm$ 2.84 & 12.16 $\pm$ 2.74 & 16.02 $\pm$ 3.32 & \textit{10.03} $\pm$ \textit{4.48} & 86.85 $\pm$ 1.48 & 88.43 $\pm$ 1.23 \\
\midrule
& Linear Modularity-Aware GAE & \textbf{28.54} $\pm$ \textbf{0.24} & 26.36 $\pm$ 0.34 & \textbf{26.38} $\pm$ \textbf{0.43} & 21.30 $\pm$ 0.59 & \textbf{84.39} $\pm$ \textbf{0.32} & \textbf{87.92} $\pm$ \textbf{0.40} \\
& Linear Standard GAE & 12.61 $\pm$ 4.61 & 6.37 $\pm$ 3.86  & 12.60 $\pm$ 4.67 & 6.21 $\pm$ 1.75 & 82.03 $\pm$ 0.32 & \textit{87.71} $\pm$ \textit{0.24} \\
\textbf{Pubmed} & Louvain & 20.06 $\pm$ 0.27 & 10.34 $\pm$ 0.99 & 16.71 $\pm$ 0.46 & 8.32 $\pm$ 0.79 & -- & -- \\
& \underline{Best other baseline:} &  &  &  &  &  & \\
& node2vec & \textit{28.52} $\pm$ \textit{1.12} & \textbf{30.63} $\pm$ \textbf{1.14} & 23.88 $\pm$ 0.54 & \textbf{25.90} $\pm$ \textbf{0.65} & 81.03 $\pm$ 0.30 & 82.33 $\pm$ 0.41 \\
\midrule
& Linear Modularity-Aware VGAE & 30.09 $\pm$ 0.63 & \textbf{29.11} $\pm$ \textbf{0.65} & \textbf{29.60} $\pm$ \textbf{0.70} & \textbf{28.54} $\pm$ \textbf{0.74} & \textit{97.10} $\pm$ \textit{0.21} & \textbf{97.21} $\pm$ \textbf{0.18} \\
\textbf{Pubmed} & Linear Standard VGAE & 29.98 $\pm$ 0.41 & \textit{29.05} $\pm$ \textit{0.20} & \textit{29.51} $\pm$ \textit{0.52} & \textit{28.50} $\pm$ \textit{0.36} & \textbf{97.12} $\pm$ \textbf{0.20} & \textit{97.20} $\pm$ \textit{0.17} \\ 
\textbf{with}  & Louvain & 20.06 $\pm$ 0.27 & 10.34 $\pm$ 0.99 & 16.71 $\pm$ 0.46 & 8.32 $\pm$ 0.79 & -- & -- \\
\textbf{features} & \underline{Best other baseline:} &  &  &  &  &  & \\
& VGAECD-OPT & \textbf{32.47} $\pm$ \textbf{0.45} & \textit{29.09} $\pm$ \textit{0.42} & \textit{29.46} $\pm$ \textit{0.52} & \textit{28.43} $\pm$ \textit{0.61} & 94.27 $\pm$ 0.33 & 94.53 $\pm$ 0.36 \\
\midrule
& Linear Modularity-Aware VGAE & \textbf{48.55} $\pm$ \textbf{0.18} & \textbf{22.21} $\pm$ \textbf{0.39} & \textbf{46.10} $\pm$ \textbf{0.29} & \textbf{20.24} $\pm$ \textbf{0.41} & \textbf{95.76} $\pm$ \textbf{0.17} & \textbf{96.31} $\pm$ \textbf{0.12} \\
& Linear Standard VGAE & 46.07 $\pm$ 0.54 & 20.01 $\pm$ 0.90 & 43.38 $\pm$ 0.37 & 18.02 $\pm$ 0.66 & \textit{95.55} $\pm$ \textit{0.22} & \textit{96.30} $\pm$ \textit{0.18} \\ 
\textbf{Cora-Large} & Louvain & 44.72 $\pm$ 0.50 & 19.46 $\pm$ 0.66 & 43.41 $\pm$ 0.52 & 19.29 $\pm$ 0.68 & -- & -- \\
& \underline{Best other baseline:}&  &  &  &  &  & \\
& DVGAE & 46.63 $\pm$ 0.56 & 20.72 $\pm$ 0.96 & 43.48 $\pm$ 0.61 & 18.45 $\pm$ 0.67 & 94.97 $\pm$ 0.23 & 95.98 $\pm$ 0.21 \\
\midrule
& Linear Modularity-Aware VGAE & \textbf{36.02} $\pm$ \textbf{0.13} & \textbf{8.12} $\pm$ \textbf{0.06} & \textbf{35.85} $\pm$ \textbf{0.20} & \textbf{8.06} $\pm$ \textbf{0.11} & \textit{82.34} $\pm$ \textit{0.38} & \textit{86.76} $\pm$ \textit{0.41} \\
& Linear Standard VGAE  & 35.01 $\pm$ 0.21 & 7.88 $\pm$ 0.15 & 30.79 $\pm$ 0.21 & 6.50 $\pm$ 0.13 & 80.11 $\pm$ 0.35 & 83.40 $\pm$ 0.36 \\ 
\textbf{SBM} & Louvain & \textit{36.00} $\pm$ \textit{0.15} & \textit{8.10} $\pm$ \textit{0.15} & \textit{35.84} $\pm$ \textit{0.18} & \textit{8.03} $\pm$ \textit{0.09} & -- & -- \\
& \underline{Best other baseline:} &  &  &  &  &  & \\
& DVGAE & \textit{35.90} $\pm$ \textit{0.18} & \textit{8.07} $\pm$ \textit{0.15} & 35.53 $\pm$ 0.23 & \textit{7.95} $\pm$ \textit{0.19} & \textbf{82.59} $\pm$ \textbf{0.36} & \textbf{87.08} $\pm$ \textbf{0.40} \\
\midrule
& GCN-Based Modularity-Aware VGAE & \textbf{21.64} $\pm$ \textbf{0.18} & \textbf{13.19} $\pm$ \textbf{0.09} & \textbf{19.10} $\pm$ \textbf{0.21} & \textbf{12.00} $\pm$ \textbf{0.17} & \textbf{85.40} $\pm$ \textbf{0.14} & \textit{86.38} $\pm$ \textit{0.15} \\
& GCN-Based Standard VGAE & 15.79 $\pm$ 0.32 & 9.75 $\pm$ 0.21 & 13.98 $\pm$ 0.35 & 8.81 $\pm$ 0.32 & \textit{85.37} $\pm$ \textit{0.12} & \textbf{86.41} $\pm$ \textbf{0.11} \\ 
\textbf{Album} & Louvain & 19.81 $\pm$ 0.19 & 12.21 $\pm$ 0.09 & 17.68 $\pm$ 0.20 & 11.02 $\pm$ 0.13 & -- & -- \\
& \underline{Best other baseline:}&  &  &  &  &  & \\
& node2vec & 20.03 $\pm$ 0.24 & 12.20 $\pm$ 0.19 & 18.34 $\pm$ 0.29 & 11.27 $\pm$ 0.28 & 83.51 $\pm$ 0.17 & 84.12 $\pm$ 0.15 \\
\bottomrule
\end{tabular}}
\end{small}
\end{center}
\end{table}

\begin{figure}[h]
\centering
\resizebox{1.03\textwidth}{!}{
  \subfigure[Linear Standard VGAE]{
  \scalebox{0.48}{\includegraphics{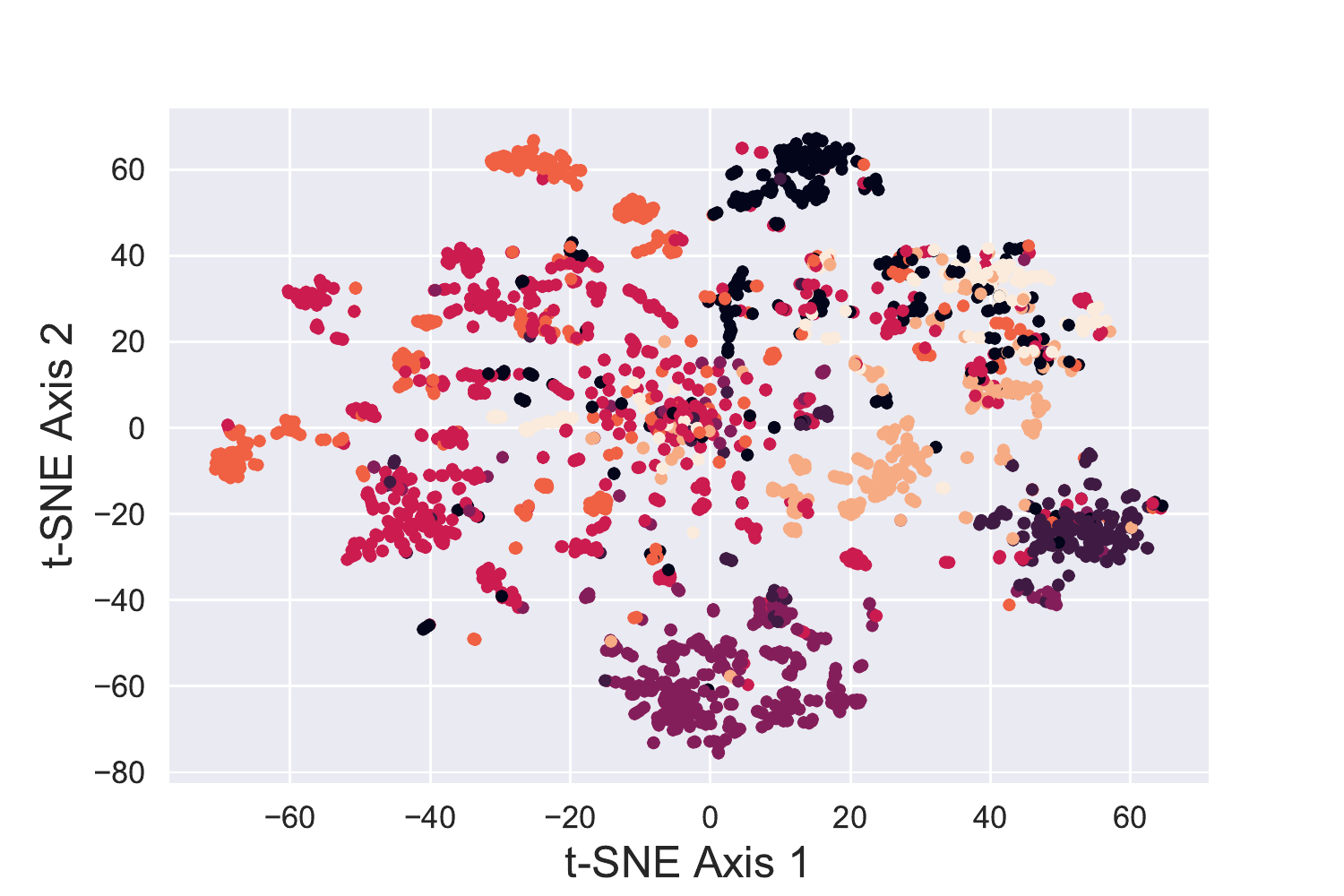}}}\subfigure[Linear Modularity-Aware VGAE]{
  \scalebox{0.48}{\includegraphics{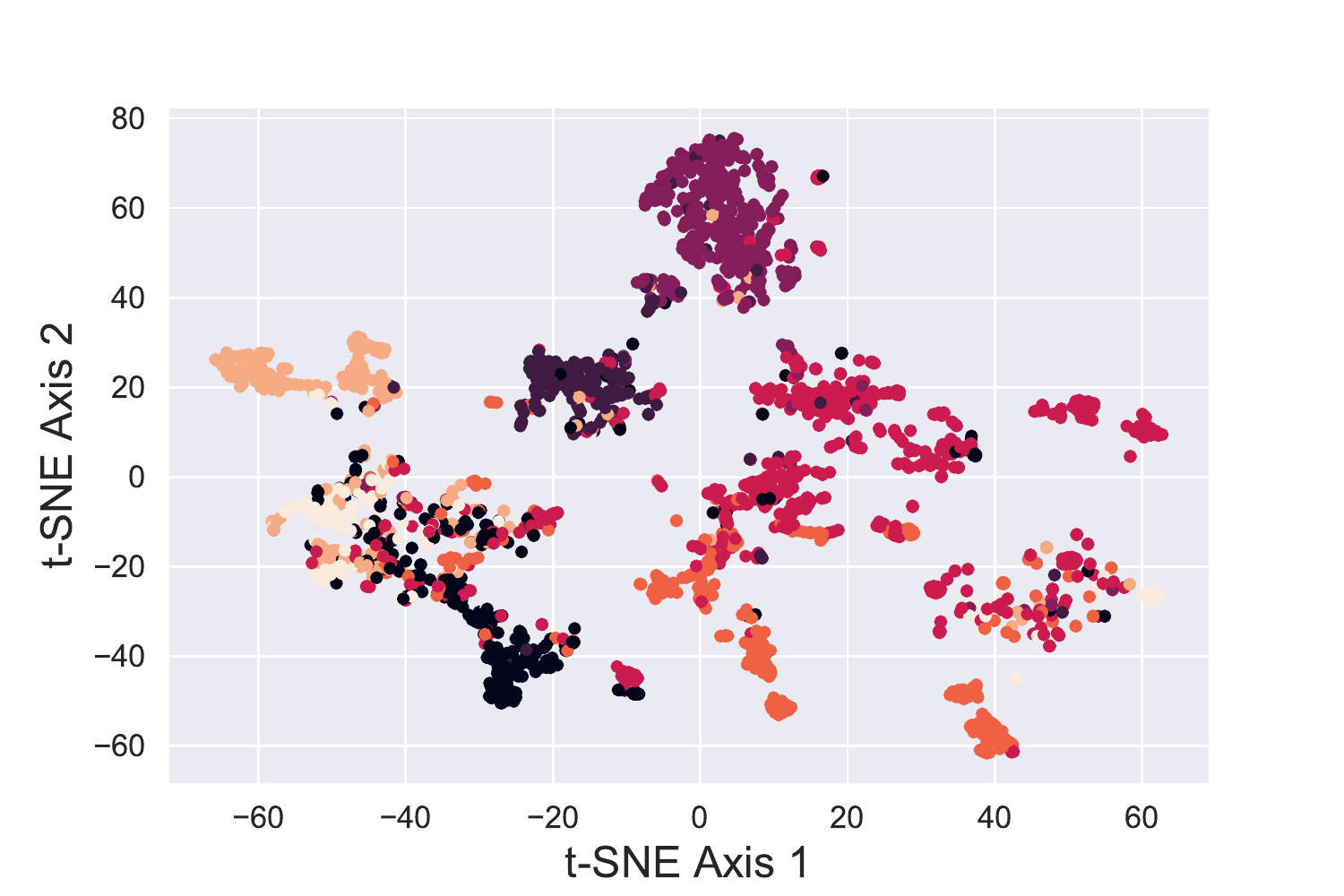}}}}
  \caption{Visualization of node embedding representations for the featureless Cora graph, learned by (a)~Standard VGAE, and (b)~Modularity-Aware VGAE, with linear encoders. The plots were obtained using the t-SNE method for high-dimensional data visualization~\cite{van2008visualizing}. Colors denote ground truth communities, that were not available during training. Although CD is not perfect (both methods return AMI scores $<$ 50\% in Table~\ref{tab:coraresults}), node embedding representations from (b) provide a more visible separation of these communities. Specifically, in Table~\ref{tab:coraresults}, using  Linear Modularity-Aware VGAE for CD leads to an increase of 9 AMI points (Task 1) to 10 AMI points (Task 2) for CD w.r.t. Linear Standard VGAE, while preserving comparable performances in LP (Task 2).}
  \label{visucora}
\end{figure}

\begin{figure}[t]
\centering
\resizebox{0.85\textwidth}{!}{
  \subfigure[Cora]{
  \scalebox{0.47}{\includegraphics{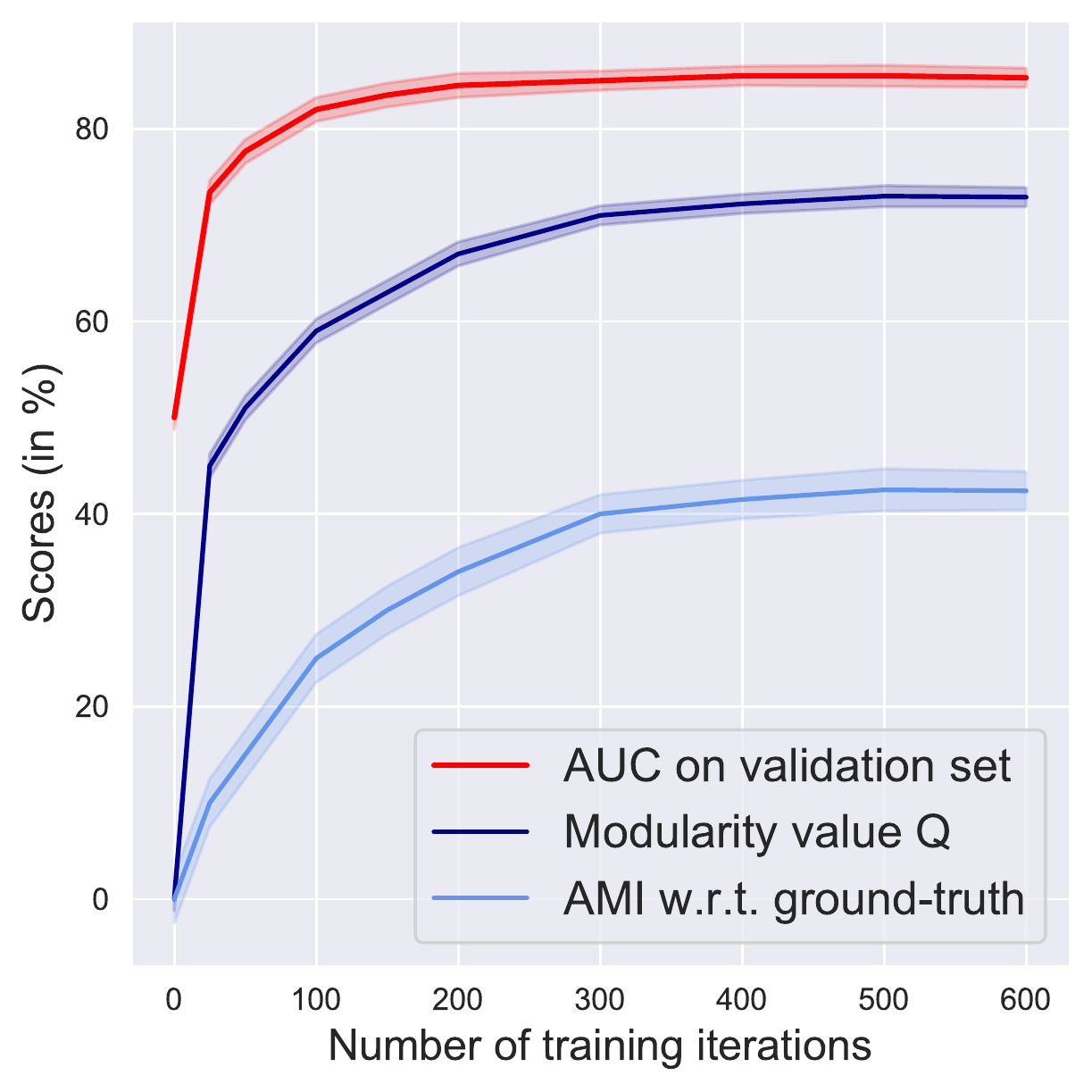}}}\subfigure[Pubmed]{
  \scalebox{0.47}{\includegraphics{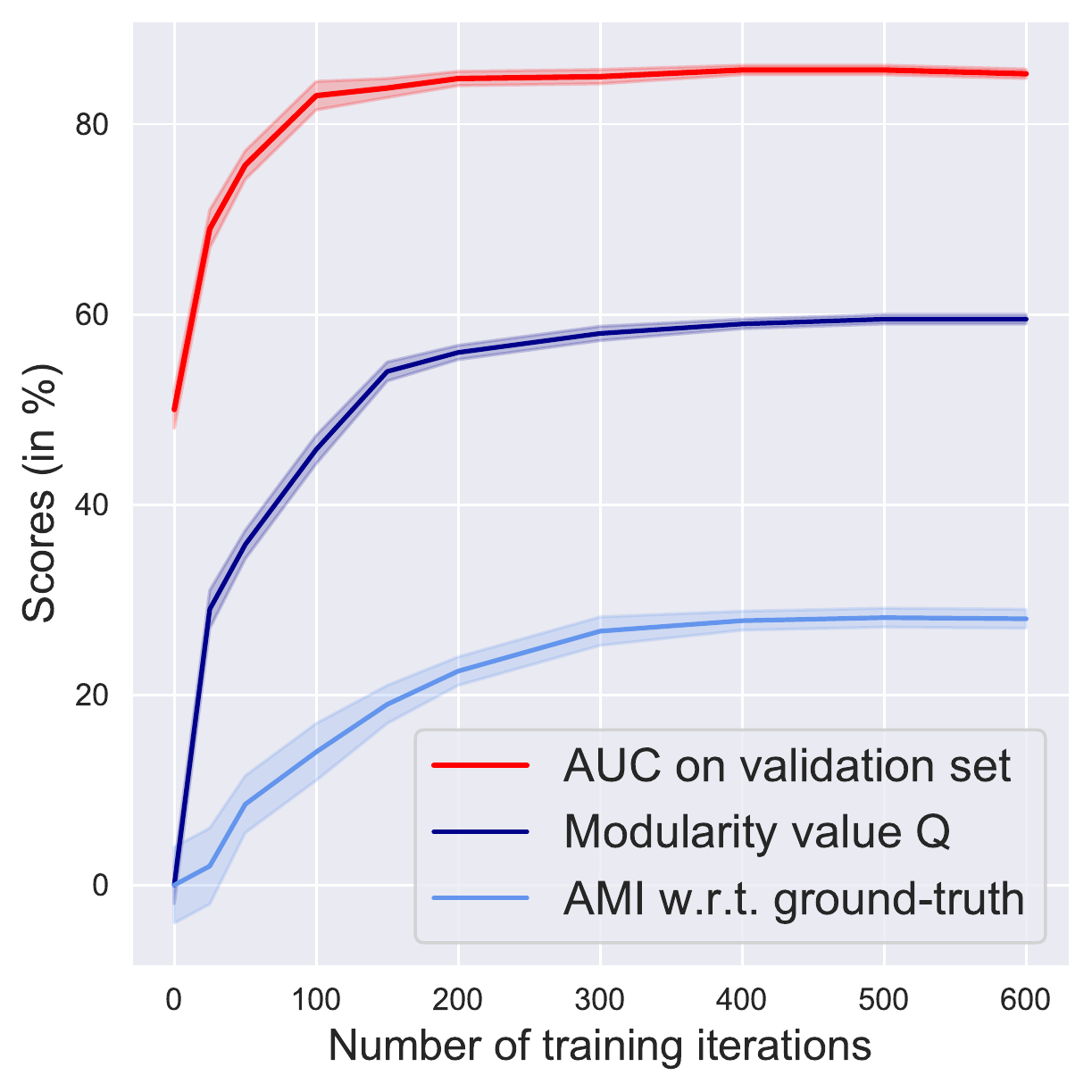}}}}
  \caption{Identification of the required number of training iterations, for Modularity-Aware VGAE with linear encoders trained on the featureless (a) Cora, and (b) Pubmed graphs. The plots report the evolution of the modularity $Q$ (\textcolor{Blue}{dark blue}) and AUC LP scores on validation sets (\textcolor{red}{red}) w.r.t. the number of training iterations in gradient ascent. By looking at the red curves only, one might choose to stop training models after 200 iterations as in \cite{kipf2016-2}, as AUC scores have almost stabilized. However, the dark blue curves emphasize that $Q$ still increases up to 400-500 training iterations for both graphs. By also using $Q$ for hyperparameter selection (as we proposed), one will therefore continue training VGAE models up to 400-500 iterations. The \textcolor{MidnightBlue}{light blue} curves confirm that such a strategy eventually leads to better AMI final scores w.r.t. ground truth communities. Note, that the light blue curves could \textit{not} be directly used for tuning, as ground truth communities are unavailable~at~training~time.}
  \label{fig:optimization}
\end{figure}

\begin{figure}[t]
\centering
\resizebox{1.0\textwidth}{!}{
  \subfigure[Cora]{
  \scalebox{0.48}{\includegraphics{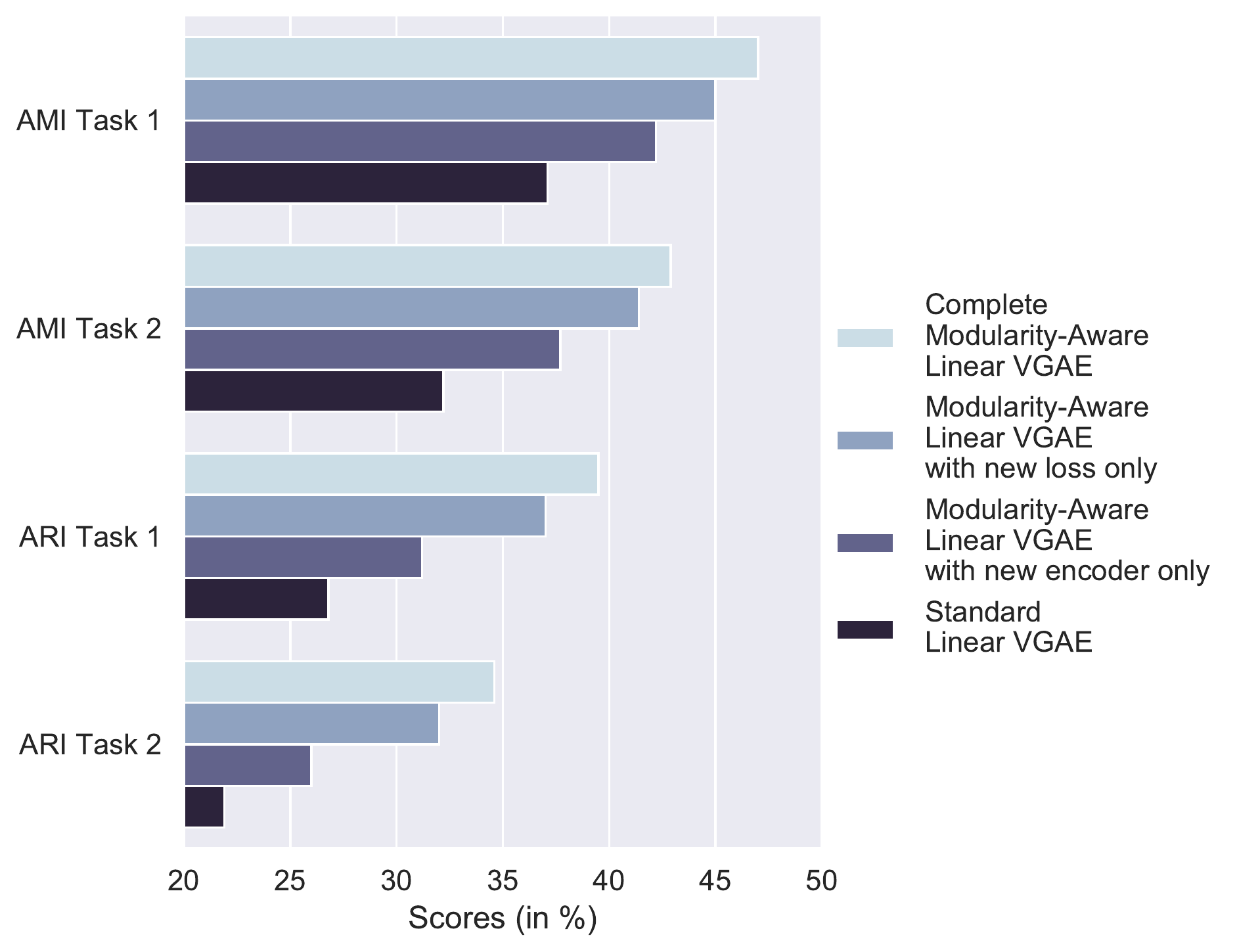}}}\subfigure[Album]{
  \scalebox{0.48}{\includegraphics{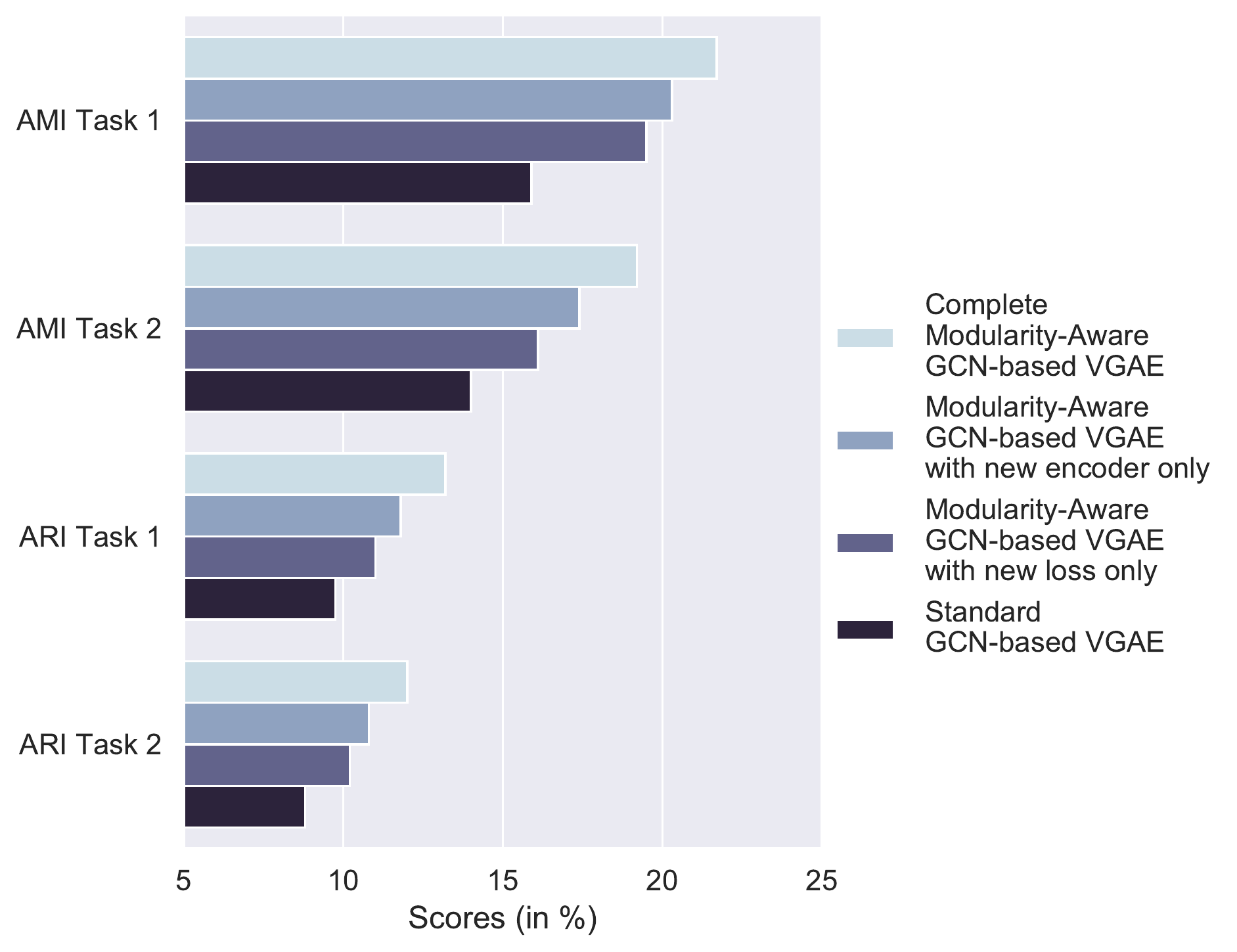}}}}
  \caption{Comparison of two ``complete'' Modularity-Aware VGAE, trained on (a) featureless Cora and (b) Album with variants of these models only leveraging our new \textit{encoder} or regularized \textit{loss} from Section~\ref{s3}. We observe that incorporating any of these two components improves CD on these graphs w.r.t. Standard VGAE. Moreover, using both components \textit{simultaneously} leads to~the~best~results.}
    \label{fig:ablation}
\end{figure}


  \begin{table}[t]
     \centering
      \caption{Normalized mutual information scores (in \%) for CD on Cora and Pubmed, \textit{with} and \textit{without} node features. \textit{Results are directly taken from the evaluation of Choong et al.}~\cite{choong2020optimizing}. This table emphasizes that, in the absence of node features, VGAECD and VGAECD-OPT bring little (to no) advantage w.r.t. standard VGAE, and remain below the Deepwalk and/or Louvain baselines. Scores of VGAECD and VGAECD-OPT significantly increase when adding features to the graph. Recall: in this table, Deepwalk and Louvain both ignore node features.}
       \vspace{0.2cm}
    \label{tab:vgaecd}
   \resizebox{0.95\textwidth}{!}{

  \begin{tabular}{c|c|cc|ccc|cc}
  \toprule
     \textbf{Dataset} & \textbf{VGAE} & \textbf{VGAECD} & \textbf{VGAECD-OPT} & \textbf{DeepWalk} & \textbf{Louvain} \\
    \midrule \midrule
    Cora \textit{without} node features & 23.84 & 28.22 & 37.35 & 37.96 & \textbf{43.36} \\

    Pubmed \textit{without} node features & 20.41 & 16.42 & 25.05 & \textbf{29.46} & 19.83 \\
\midrule
    Cora \textit{with} node features& 31.73 & 50.72 & \textbf{54.37} & 37.96 & 43.36\\
    Pubmed \textit{with} node features & 19.81 & 32.53 & \textbf{35.52} & 29.46 & 19.83\\
    
    \bottomrule
  \end{tabular}}

\end{table}

\end{document}